%% file: sepmf_iclr.tex
\documentclass{article}
\usepackage{iclr2021_conference,times}
\input{math_commands.tex}

\usepackage{hyperref}
\usepackage{url}

\usepackage[utf8]{inputenc} 
\usepackage[T1]{fontenc}    
\usepackage{hyperref}       
\usepackage{url}            
\usepackage{booktabs}       
\usepackage{amsfonts}       
\usepackage{nicefrac}       
\usepackage{microtype}      
\usepackage{amsmath}
\usepackage{amssymb}
\usepackage{mathtools} 

\usepackage{amsthm}

\theoremstyle{definition}
\newtheorem{definition}{Definition}

\newcommand{\bA}{\mathbf{A}}
\newcommand{\bY}{\mathbf{Y}}

\newcommand{\btheta}{\boldsymbol{\theta}}

\newcommand{\dd}{\textrm{d}}
\usepackage{mathtools}
\usepackage{graphicx}
\graphicspath{{../figures/}}
\usepackage{subcaption}
\usepackage{color,soul}

\author{Joshua C. Chang \\
    National Institutes of Health\\
    Clinical Center    \&  \href{https://mederrata.com}{team {med$\varepsilon_{\text{rrata}}$}} \\
     \href{mailto:josh@mederrata.com}{\tt  josh@mederrata.com} 
\And
    Patrick A. Fletcher \\
    National Institutes of Health\\
    NIDDK 
     \&  team {med$\varepsilon_{\text{rrata}}$}  \\
      \href{mailto:patrick@mederrata.com}{
      \texttt{patrick@mederrata.com}}
    \And 
Jungmin Han \\
    National Institutes of Health\\
    NIDDK \&
  team {med$\varepsilon_{\text{rrata}}$}  \\
  \href{mailto:jungmin@mederrata.com}{\tt jungmin@mederrata.com}
      \And
    Ted L. Chang \\ 
    team {med$\varepsilon_{\text{rrata}}$}\\ \href{mailto:ted@mederrata.com}{\tt ted@mederrata.com} 
     \And Shashaank Vattikuti  \\
        National Institutes of Health\\ NIDDK
 \& team {med$\varepsilon_{\text{rrata}}$} \\ 
 \href{mailto:shashaank@mederrata.com}{\tt shashaank@mederrata.com} 
\And
    Bart Desmet \\
    National Institutes of Health\\ Clinical Center \\
    \href{mailto:bart.desmet@nih.gov}{\tt bart.desmet@nih.gov} 
    \And 
    Ayah Zirikly  \\
    National Institutes of Health\\ Clinical Center \\ 
     \href{mailto:ayah.zirikly@gmail.com}{\tt ayah.zirikly@gmail.com} 
    
  \And Carson C. Chow  \\
   National Institutes of Health\\ NIDDK \\
 \href{mailto:carsonc@niddk.nih.gov }{\tt carsonc@nih.gov} 
}

\title{Sparse encoding for more-interpretable feature-selecting representations in probabilistic matrix factorization}

\iclrfinalcopy 

\begin{document}
\maketitle

\begin{abstract} 

Dimensionality reduction methods for count data are critical to a wide range of applications in medical informatics and other fields where model interpretability is paramount.
For such data, hierarchical Poisson matrix factorization (HPF) and other sparse probabilistic non-negative matrix factorization (NMF) methods are considered to be interpretable generative models.
They consist of sparse transformations for decoding their learned representations into predictions. 
However, sparsity in representation decoding does not necessarily imply sparsity in the encoding of representations from the original data features. 
HPF is often incorrectly interpreted in the literature as if it possesses encoder sparsity.
The distinction between decoder sparsity and encoder sparsity is subtle but important.
Due to the lack of encoder sparsity, HPF does not possess the column-clustering property of classical NMF -- the factor loading matrix does not sufficiently define how each factor is formed from the original features.
We address this deficiency by self-consistently enforcing encoder sparsity, using a generalized additive model  (GAM), 
thereby allowing one to relate each representation coordinate to a subset of the original data features. 
In doing so, the method also gains the ability to perform feature selection.
We demonstrate our method on simulated data and give an example of how encoder sparsity is of practical use in a concrete application of representing inpatient comorbidities in Medicare patients.

\end{abstract}

\section{Introduction}

For many inverse problems, such as those found in healthcare, model interpretability is paramount. 
Building interpretable high-performing solutions is technically challenging and requires flexible frameworks.
A general approach to these and other problems is to structure solutions into pipelines.
By ensuring interpretability of each step in such a pipeline, one can achieve interpretability of  the overall larger model.

Data sources for natural language processing, medical informatics, and bioinformatics are often high-dimensional count matrices.
A common first step in modeling high-dimensional data sets is to use dimensionality reduction to find tractable data representations (also called factors or embeddings), that are then fed into downstream analyses. 
Our goal is to develop a dimension reduction scheme for count matrices such that the reduced representation has an innate interpretation in terms of the original data features.

\subsection{Interpretability versus explainability}

We seek latent data representations that are not only post-hoc explainable~\citep{laugelDangersPosthocInterpretability2019,caruanaIntelligibleExplainableMachine2020}, but also intrinsically interpretable~\citep{rudinStopExplainingBlack2019}.
Our definition of intrinsic interpretability requires that it is clear how the model outputs are computed from the original variables, and that the latent variables and interactions within the model are  meaningful.

Post-hoc explanations are based on subjective examination of a solution through the lens of subject-matter expertise.
For black-box models that lack intrinsic interpretability, these explanations are produced using simpler approximating models (typically local linear regressions).
In addition to the explanations being inexact, they can also be misleading~\citep{laugelDangersPosthocInterpretability2019}.
Examples of black-box models that rely on approximation techniques for explainability are found throughout the deep learning literature. 

\subsection{Disentangled autoencoders}
Disentangled variational autoencoders~\citep{higginsBetaVAELearningBasic2016,tomczakVAEVampPrior2017,dengFactorizedVariationalAutoencoders2017} are deep learning models that are inherently mindful of post-hoc model explainability. Like other autoencoders, these models are encoder-decoder structured (see Definitions~\ref{def:encoder} and \ref{def:decoder}), where the encoder generates dimensionally reduced representations.

\theoremstyle{definition}
\begin{definition}
The \textbf{encoder} transformation maps input data features to latent representations
\label{def:encoder}
\end{definition}

\theoremstyle{definition}
\begin{definition}
The \textbf{decoder} transformation maps latent representations to predictions
\label{def:decoder}
\end{definition}

Disentangled autoencoders use a combination of  penalties~\citep{higginsBetaVAELearningBasic2016,hoffmanBetaVAEImplicit2017} 
and structural constraints~\citep{ainsworthInterpretableVAEsNonlinear2018} to encourage statistical independence in representations, facilitating explanation.
These methods arose in computer vision and have demonstrated empirical utility in producing nonlinear factor models where the factors are conceptually sensible.
Yet, due to the black-box nature of deep learning, explanations for how the factors are generated from the data, using local saliency maps for instance, are unreliable or imprecise~\citep{laugelDangersPosthocInterpretability2019,slackFoolingLIMESHAP2020,arunAssessingTrustworthinessSaliency2020}.
In imaging applications, where the features are raw pixels, this type of interpretability is unnecessary. 
However, when modeling structured data problems, one often wishes to learn the effects of the individual data features.

\subsection{Probabilistic matrix factorization}

Probabilistic matrix factorization methods are related to autoencoders~\citep{mnihProbabilisticMatrixFactorization2008}.
These methods are often presented in the context of recommender systems.
In these cases, rows of the input matrix are attributed to users, and columns (features) are attributed to items.
Probabilistic matrix factorization methods are bi-linear in item- and user-specific effects, de-convolving them in a manner similar to item response theory~\citep{changProbabilisticallyautoencodedHorseshoedisentangledMultidomain2019}.
In applications with non-negative data, non-negative sparse matrix factorization methods further improve on interpretability by computing predictions using only additive terms~\citep{leeLearningPartsObjects1999}.

For count matrices, \citet{gopalanScalableRecommendationPoisson2014} introduced hierarchical Poisson matrix factorization (HPF).
Suppose $\mathbf{Y}=(y_{ui})$ is a $U\times I$ matrix of non-negative integers, where each row corresponds to a \emph{user} and each column corresponds to an \emph{item} (feature).
Adopting their notation, \citet{gopalanScalableRecommendationPoisson2014} formulated their model as
\begin{align}
    y_{ui}\vert \boldsymbol{\Theta}, \mathbf{B}\sim \textrm{Poisson} \left(\sum_k\theta_{uk}\beta_{ki} \right)  \qquad
    \begin{aligned}
        \theta_{uk}\vert \xi_u, a&\sim\textrm{Gamma}\left( a, \xi_u\right) \\
        \beta_{ki}\vert \eta_i, c&\sim\textrm{Gamma}\left( c, \eta_i\right),
    \end{aligned}
    \label{eq:hpf}
\end{align}
where $\boldsymbol\Theta = (\theta_{uk})$ is a \ $U\times K$ matrix, and $\mathbf{B} = (\beta_{ki})$ is the representation decoder matrix.  Additional priors $\eta_i\sim\textrm{Gamma}(c', c'/d')$ and $\xi_u\sim\textrm{Gamma}(a', a'/b')$ model item and user-specific variability in the dataset, and $a',b',c',d'\in\mathbb{R}^+$ are  hyper-parameters. 
The row vector $\boldsymbol\theta_u = (\theta_{u1},\ldots, \theta_{uK})$ constitutes a $K$-dimensional \emph{representation} of the user,
and the matrix $\mathbf{B} = (\beta_{ki})$ decodes the representation into predictions on the user's counts. 

\begin{figure}
    \centering
    \includegraphics[width=\textwidth]{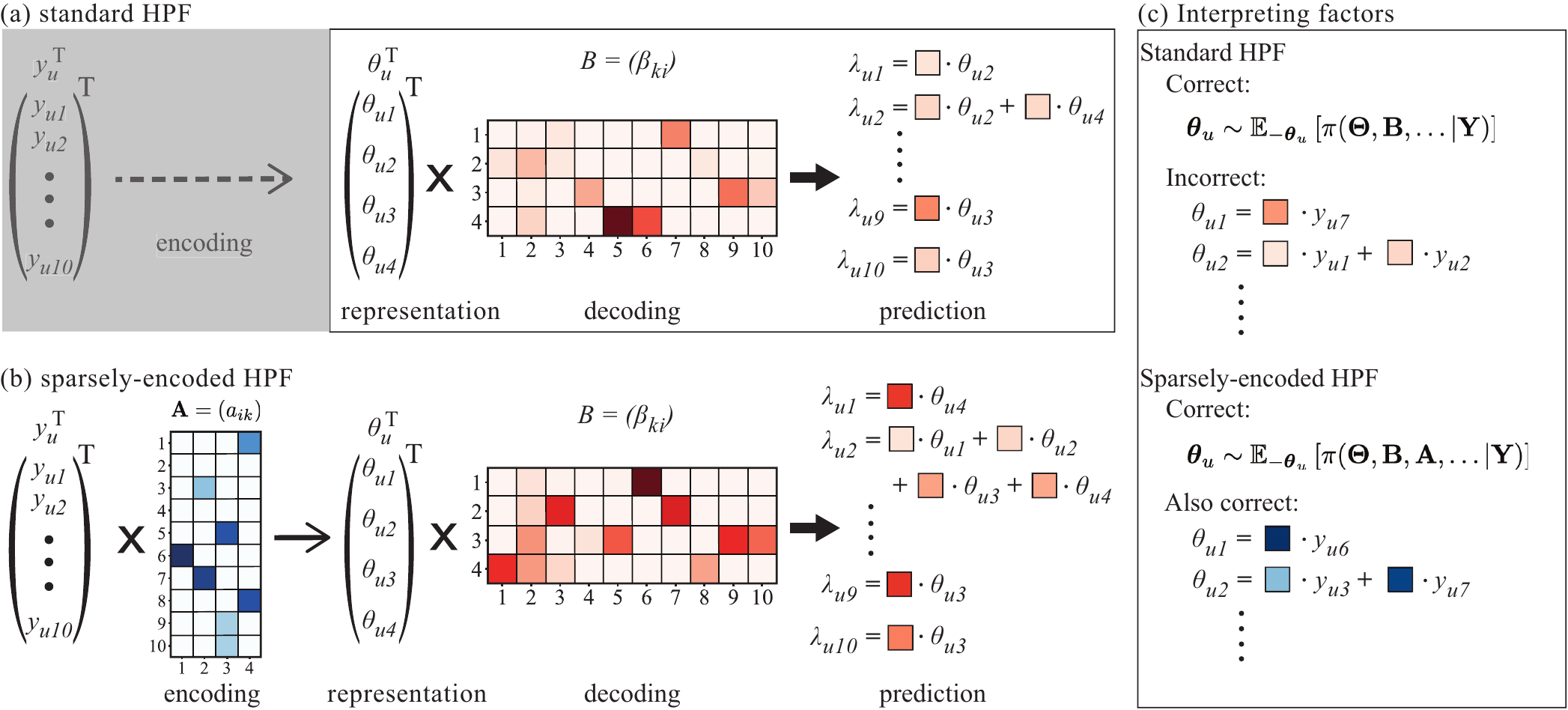}
    \caption{\textbf{Interpreting hierarchical sparse probabilistic matrix factorization (HPF).}  \textbf{(a) Standard HPF:} Rates $\lambda_{ui}$ for Poisson-distributed predictions are sparse linear combinations of the learned representation as defined by the decoding matrix; this matrix does not define how representations are derived from the input data.  \textbf{(b) Sparsely-encoded HPF (proposed method):} the mapping from input data to representation is given explicitly by a sparse encoding matrix. \textbf{(c) Interpreting representations:} It is tempting but misleading to read the decoding matrices row-wise in determining the feature subsets that contribute to forming a representation coordinate. Representations $\boldsymbol{\theta}_u$ are computed by inferring the statistics of an associated joint posterior distribution $\pi(\ldots\vert \mathbf{Y})$ -- the sets of non-sparse entries in rows of the decoding matrices do not necessarily correspond to feature sets that determine the representation components. However, for sparsely-encoded HPF, the representations are explicit functions of subsets of features. }
    \label{fig:interpretability}
\end{figure}

In HPF, the gamma priors  on the decoder matrix  $\mathbf{B} = (\beta_{ki})$ enforce non-negativity. Because the gamma distribution can have density at zero, these priors also allow for sparsity where only a few of the entries are far from zero.
Sparsity, non-negativity, and the simple bi-linear structure of the likelihood in HPF combine to yield a simple interpretation of the model:
in HPF, a predictive density for each matrix element is formed using a linear combination of a subset of representation elements, where the elements of $\mathbf{B}$ determine the relative additive contributions of each of the elements (Fig.~\ref{fig:interpretability}a). However, the composition of each latent factor in terms of the original items is not explicitly determined but arises from Bayesian inference (Fig.~\ref{fig:interpretability}c).

\subsection{Limitations of HPF}\label{sec:limitations}
Classical non-negative matrix factorization (NMF) is often touted for having a column-clustering property~\citep{dingEquivalenceNonnegativeMatrix2005},
where data features are grouped into coherent factors.
The standard HPF of Eq.~\ref{eq:hpf} lacks this property.
In HPF, while each prediction is a linear combination of a subset of factors,  each factor is not necessarily a linear combination of a subset of features (depicted in Fig.~\ref{fig:interpretability}c).

The transformation matrix $\mathbf{B}$ defines a decoder (Def.~\ref{def:decoder}) like a classic autoencoder. A corresponding encoding  transformation (Def.~\ref{def:encoder}) does not explicitly appear in the formulation of Eq.~\ref{eq:hpf}.
Determining the composition of factors is not simply a matter of reading the decoding matrix row-wise.
Mathematically, sparsity in decoding does not imply sparsity in encoding analogous to how pseudo-inverses of sparse matrices are not necessarily sparse.

HPF is also unable to perform feature selection. By Eq.~\ref{eq:hpf}, predictions are formed by weighting representations using values from columns of the decoding matrix -- a feature's corresponding terms in the decoding matrix will be near zero if and only if that feature's mean is near zero. Exclusion of a feature column from the decoding matrix yields no information on whether that  feature plays a part in generating representations.
Also, this deficiency can cause HPF to erroneously imply structure when none is present, as demonstrated in Fig.~\ref{fig:hpfrec_sims}a) on a factorization of pure Poisson noise.

\subsection{Our contributions}
\label{sec:contributions}

We propose a method to self-consistently constrain HPF so that its corresponding encoding transformation is explicit. In doing so, we improve interpretability of HPF, and give it the ability to perform feature selection. 
Constraining HPF in this manner also makes it more suitable to training with large datasets
because the representation matrix does not need to be stored in memory.
Using a medical claims case study, we demonstrate how our method facilitates reparameterization of decision rules in representation space into corresponding rules on the original data features.


\section{Methods}

In this section we describe our extension to HPF that resolves the issues mentioned in Section~\ref{sec:limitations}. Our method yields representations that have explicit sparse dependence on relevant features in the input data.

\subsection{Improving HPF by constraining it}
\label{sec:model}
Our augmented HPF model takes the form
\begin{align}
    y_{ui}\vert \boldsymbol{\Theta}, \mathbf{B},\boldsymbol{\varphi}\sim \textrm{Poisson} \left(f_i\left(\sum_k \theta_{uk}\beta_{ki}\right) + \varphi_i \right)  \quad
    \begin{aligned}[c]
        \theta_{uk}\vert \bA, \mathbf{y}_u, \xi_u &= \xi_u\sum_i g_i(y_{ui}) \alpha_{ik}    \\ 
        \
        \beta_{ki}&\sim \textrm{Normal}^+(0, 1/2K),
        \end{aligned}
    \label{eq:likelihood}
\end{align}
where prior distributions for the model parameters are defined later in this section.

The key point is that the encoder function (that computes $\theta_{uk}$) is an explicit function of the input data, formulated using a generalized additive model (GAM)~\citep{rigbyGeneralizedAdditiveModels2005,hastieGeneralizedAdditiveModels1987,kleinBayesianGeneralizedAdditive2015}. 
The encoding matrix  $\mathbf{A}=(\alpha_{ik})$ controls how features map into the representation.

To allow the model to perform automatic feature selection, we also incorporate a non-negative item-specific gain term $\varphi_i$ as a background Poisson rate for item~$i$ that is intended to be independent of the factor model.
We also slightly generalize the likelihood of Eq.~\ref{eq:hpf} by giving each feature an associated link function $f_i$, which models nonlinearities without sacrificing interpretability.

The distributions of the parameters of the encoder are learned self-consistently with other model parameters.
In the process, one is training not only the generative model, but also the subsequent Bayesian inference of mapping data to representation by learning the statistics of the posterior distribution,
\begin{equation}
\btheta_u \vert \mathbf{y}_u, \bY \sim  \iint \pi(\btheta_u , \mathbf{B},\varphi \vert \mathbf{y}_u, \bY)\dd\mathbf{B}\dd\varphi,
\end{equation}
where the generative process has been marginalized.
In short, the model of Eq.~\ref{eq:likelihood} uses the marginal posterior distribution of the encoding matrix $\bA$ to reparameterize this Bayesian inference. 
Doing so amortizes this inference, making it trivial to apply the model to new data in order to compute new representations.
It also allows us to impose desirable constraints on the representations themselves.

In the original HPF, the parameters $\xi_u$ are used to account for variability in user activity (row sums).
Similarly, the $\eta_i$ parameters account for variability in item popularity (in column sums). 
To simplify the method, we pre-set these parameters (based on some training data) to $\xi_u=1$ and
$
    \eta_i = \frac{1}{U}\sum_u y_{ui} $,
where $\eta_i$ is absorbed into the function $f_i.$
Doing so de-scales the encoder parameters $\alpha_{ik}$ so we can generalize weakly-informative and other scale-dependent priors within the model to disparate datasets, as is common in preprocessing for Bayesian statistical inference problems~\citep{gelmanPriorCanOften2017}.
One may also model over-dispersed data  using $\xi_u = \nicefrac{U\sum_i y_{ui}}{ \sum_u \sum_i y_{ui}}$, to account for  document-size variability.

Eq.~\ref{eq:likelihood} implies that we no longer promote sparsity in either $\theta_{uk}$ or $\beta_{ki}.$
Instead, we use sparsity to aid feature selection.
We encourage the elements $\alpha_{ik}$ and $\varphi_i$ to be mutually exclusive by using the decomposition
\begin{equation}
    \begin{aligned}
            \alpha_{ik} &= u_{ik}\frac{s_i^+}{s_i^+ + s_i^-} \\
        \varphi_i \vert \eta_i,w_i, s_i^\pm &= \eta_iw_i \frac{s_i^-}{s_i^+ + s_i^-}
    \end{aligned} \qquad\qquad
    \begin{aligned}
    [s_i^+ \ s_i^-]^\intercal &\sim \textrm{Horseshoe}^+(1,1) \\
    [u_{1k}\ u_{2k} \ \ldots \  u_{Ik}]^\intercal&\sim \textrm{Horseshoe}^+(1, \nicefrac{1}{\sqrt{UI}}) \\
    w_i  &\sim \textrm{Normal}^+(0, 10),
    \end{aligned}
\end{equation}
where the non-negative version of the Horseshoe$^+$ prior ~\citep{carvalhoHandlingSparsityHorseshoe2009,carvalhoHorseshoeEstimatorSparse2010,polsonHalfCauchyPriorGlobal2011} is the hierarchical Bayesian model
\begin{equation}
 \mathbf{x}\sim \textrm{Horseshoe}^+(\lambda_0, \tau_0) \qquad \Longleftrightarrow \quad \begin{aligned}
 x_j \vert \lambda_j, \tau&\sim \textrm{Normal}^+(0, \lambda_j\tau) \\ 
 \lambda_j &\sim \textrm{Cauchy}^+(0, \lambda_0) \\
\tau &\sim\textrm{Cauchy}^+(0, {\tau_0}).
 \end{aligned}
 \label{eq:horseshoe}
\end{equation}
This concentrates marginal distributions of vector components near zero. 
Additionally, it minimally shrinks large components, resulting in lower bias compared to lasso and other alternatives~\citep{bhadraHorseshoeEstimatorUltraSparse2015,bhadraDefaultBayesianAnalysis2015,bhadraLassoMeetsHorseshoe2019,piironenSparsityInformationRegularization2017}.
The horseshoe has previously been applied in other factorization methods, including autoencoders~\citep{ghoshModelSelectionBayesian2017a} and item response theory~\citep{changProbabilisticallyautoencodedHorseshoedisentangledMultidomain2019},
but not to probabilistic matrix factorization.

Applied to the parameters $s_i^\pm$, sparsity discourages variables that load into the factor model from leaking into the corresponding background rate term $\varphi_i.$
Conversely, variables that load into  $\varphi_i$ are discouraged from appearing in $\alpha_{ik}$.

%
Finally, as is often done in variational autoencoders, 
we regularize the representation by placing unit half normal priors on its components
\begin{align}
\theta_{uk}\vert g, \bY, \bA = \xi_u\sum_i g_i(y_{ui})\alpha_{ik}&\sim\textrm{Normal}_{\theta_{uk}}^+(0,1). \label{eq:theta_constraint}
\end{align}
The choices of the encoding function $f_i$ and decoding functions $g_i$ are application-specific, and may be learned~\citep{rigbyGeneralizedAdditiveModels2005}.
So as not to distract from our focus on improving the interpretability of pre-existing matrix factorization approaches, we fix these functions here.
In standard Poisson matrix factorization approaches,  
$
    f_i(x) = g_i(x) = x,\ \forall{i}. 
$
Equivalently, we choose to rescale the inputs so that
\begin{equation}\label{eq:standard_link}
    f_i(x) = \eta_i x \qquad \textrm{and}\qquad g_i(x) = f_i^{-1}(x) = \nicefrac{x}{\eta_i}.
\end{equation}
Another choice for these functions can be motivated by Poisson regression with a logarithmic link function, 
where
$
    f_i(x) = e^{\eta_i x } -1, \ 
$
and
$g_i(x) = f_i^{-1}(x)= \log\left( \nicefrac{x}{\eta_i}+1\right)$. 
%
For maximum interpretability, restricting $f_i$ and $g_i$ to monotonically increasing functions where $g_i(0)=f_i(0)=0$ results in order-preserving representations that are zero when the corresponding feature counts are zero.

\subsection{Interpreting representations and derived quantities}

In constraining the encoder mapping using the generalized additive model of Eq.~\ref{eq:theta_constraint}, we regain the column-clustering property of classical non-negative matrix factorization methods: each representation component is explicitly determined from a well-defined subset (cluster) of the data features.
In Fig.~\ref{fig:interpretability}c), we demonstrate how the encoding matrix can be read to determine the composition of the factors.
Consequently, decision rules over the representation can be easily expressed as decision rules over the original features,
\begin{equation}
   \theta_{uk} \in (a,b)  \Longleftrightarrow \sum_{j\in\Omega_k} g_j (y_{uj} ) \alpha_{jk} \in (\nicefrac{a}{\xi_u}, \nicefrac{b}{\xi_u}),
   \label{eq:reparameterization}
\end{equation}
where $\Omega_k$ is the subset of features that determines factor $k$.
As we will demonstrate in our main case study, Eq.~\ref{eq:reparameterization} is useful for inverting clustering rules defined over the representations.

\subsection{Inference}

The model of Eq.~\ref{eq:likelihood} is a generalized linear factor model that we have mathematically related to a probabilistic autoencoder.
When augmenting HPF with explicit encoder inference, as we have done, one obtains a probabilistic autoencoder.
This suggests that previous work can serve as a guide for training, 
especially work done on using the horseshoe prior in Bayesian neural networks~\citep{ghoshModelSelectionBayesian2017,ghoshStructuredVariationalLearning2018,louizosBayesianCompressionDeep2017}.

In particular, \citet{ghoshStructuredVariationalLearning2018} investigated structured variational approximations of inference of Bayesian neural networks that use the horseshoe prior and found them to have similar predictive power as mean-field variational approximations.
The disadvantage of structured approximations is the extra computational cost of inferring covariance matrices.
For these reasons, we focus on mean-field black-box variational inference, using~\citet{ghoshStructuredVariationalLearning2018} as a guide, noting consistency of their scheme with other works that have investigated variational inference on problems using the horseshoe prior~\citep{wandMeanFieldVariational2011,louizosBayesianCompressionDeep2017}.

As in~\citet{ghoshModelSelectionBayesian2017,ghoshStructuredVariationalLearning2018,changProbabilisticallyautoencodedHorseshoedisentangledMultidomain2019}, for numerical stability, we reparameterize the Cauchy distributions in terms of the auxiliary inverse Gamma representation~\citep{makalicSimpleSamplerHorseshoe2016},
\begin{equation}
\begin{split}
    x\sim\textrm{Cauchy}^+(0, \sigma)
\end{split}
      \qquad\Longleftrightarrow
 \qquad
      \begin{array}{l}
          x^2\sim \text{Inverse-Gamma}\left(\nicefrac{1}{2},\ \nicefrac{1}{\lambda}\right) \\
          \lambda \sim \text{Inverse-Gamma}\left(\nicefrac{1}{2},\ \nicefrac{1}{\sigma^2}\right).
      \end{array}
      \label{eq:IGamma_parameterization}
\end{equation}

We perform approximate Bayesian inference using fully-factorized mean-field Automatic Differentiation Variational Inference (ADVI)~\citep{kucukelbirAutomaticDifferentiationVariational2017}.
For all matrix elements, we utilized softplus-transformed Gaussians, and coupled these to inverse-Gamma distributions for the scale parameters, as investigated in~\citet{wandMeanFieldVariational2011}.
For our use cases, we implemented a minibatch training regimen common to machine learning,
with stepping given by the Adam optimizer~\citep{kingmaAdamMethodStochastic2017} combined with the Lookahead algorithm~\citep{zhangLookaheadOptimizerSteps2019} for stabilization.

Bayesian sparsity methods concentrate marginal distributions for parameters near zero.
To further refine the Bayesian parameter densities so that some parameters are identically zero in distribution, one may use projection-based sparsification~\citep{piironenSparsityInformationRegularization2017}.
Finally, one can assess predictive power without model refitting using approximate leave-one-out cross validation (LOO) using the Widely Applicable Information Criterion (WAIC)~\citep{watanabeAsymptoticEquivalenceBayes2010,gelmanUnderstandingPredictiveInformation2014,piironenComparisonBayesianPredictive2017,vehtariPracticalBayesianModel2017,changPredictiveBayesianSelection2019}
  or Pareto smoothed importance sampling LOO (PSIS-LOO)~\citep{vehtariPracticalBayesianModel2017}.

\section{Experiments}

We implemented the method in {tensorflow-probability}~\citep{dillonTensorFlowDistributions2017}, 
modifying the inference routines for implementing our variational approximation. 
Our implementation can be found at \href{https://www.github.com/mederrata/spmf}{\texttt{github:mederrata/spmf}}, along with notebooks reproducing our simulation results.
We present here simulation results and an application to medical claims data.  

\subsection{Simulations}

To demonstrate the properties of our method, we factorized synthetic datasets of: a) completely random noise with no underlying structure, b) a mixture of random noise and linear structure, and c) a mixture of random noise and nonlinear structure.

\begin{figure}[!h]
    \centering
    \includegraphics[width=\linewidth]{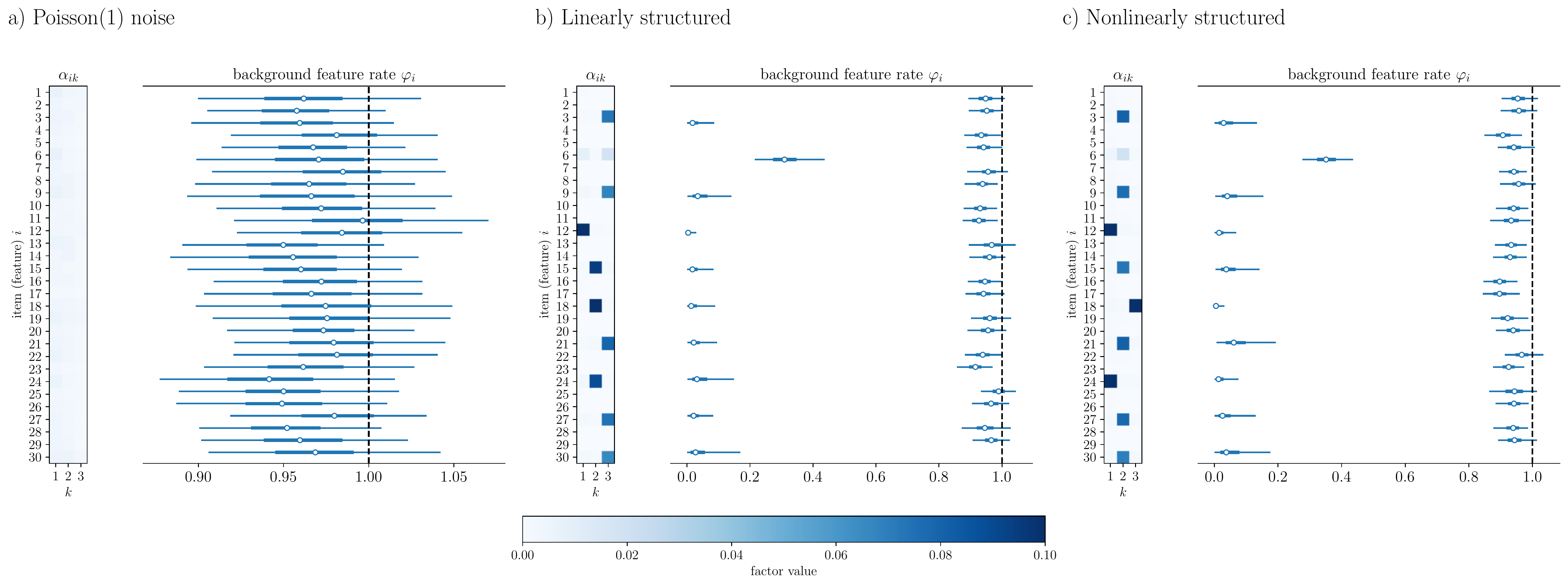}
\caption{\textbf{Factorization of simulated datasets}. The (mean) effective encoding matrix $\bA = (\alpha_{ik})$ for each factor process, placed on a common color scale, and the posterior distribution of the background process rate $\varphi_i$ by item for a)  Poisson($1$) noise,  where there is no relationship between the features, b) linear factor model where every third variable is generated from a dense factor model and the other variables are Poisson($1$) noise, c) nonlinear factor model where every third variable is generated from a dense nonlinear factor model and the other variables are Poisson($1$) noise. See Fig.~\ref{fig:hpfrec_sims} for standard HPF on these datasets for comparison.}
    \label{fig:simulation_factorization}
\end{figure}

For a), we sampled a $50,000\times 30$ Poisson($1$) random matrix.
Figure~\ref{fig:simulation_factorization}a) shows the inferred mean encoder matrix $\mathbf{A}$ along with the posterior distributions for each of the background components $\varphi_i$. 
We see that all features are excluded from the  encoding matrix, showing up instead as background noise.

Next, we created a test system where there is underlying linear structure mixed with noise. For this system, we again used $I=30$ features and put every third feature into a dense system by generating a random $10\times 10$ decoding matrix $\mathbf{B}$, sampling representations from a non-negative truncated normal distribution, and sampling counts according to the generative process of Eq.~\ref{eq:hpf}. For the remaining features, we used Poisson($1$) noise. After simulating $50,000$ records by this process, we performed factorization again into $K=3$ dimensions. The results of this factorization are shown in Figure~\ref{fig:simulation_factorization}b), where it is clear that every third feature falls into the overall factor model and the remaining features show up as background noise.

As an example of factorization under model mismatch, we generated random data with underlying nonlinear structure. Here again, we used every third feature, simulating $\mathbf{B}$ and $\boldsymbol\Theta$ as before. However, we simulated counts for these features using the model
$
y_{ui} \sim \textrm{Poisson}\left(\left(\sum_{k}\nicefrac{\theta_{uk}\beta_{ki}}{2}\right)\exp\left(-\sum_{k}\nicefrac{\theta_{uk}\beta_{ki}}{2}\right) + \left(\sum_{k}\nicefrac{\theta_{uk}\beta_{ki}}{2}\right)^{2} \right).
$
 Factorization of this dataset is shown in  Figure~\ref{fig:simulation_factorization}c). 
Again, it is clear that every third feature falls into the overall factor model and the remaining features are load into the background process with rates near $1$, indicating that even when the model is mis-specified, it can successfully separate structure from noise. 
In Supplemental Fig.~\ref{fig:log_sim_factorization}, we present the same factorizations while using the logarithmic link function described in Sec.~\ref{sec:model}, demonstrating robustness to mis-specification of the link functions as measured using the WAIC.

\begin{figure}
    \centering
    \includegraphics[width=0.6\textwidth]{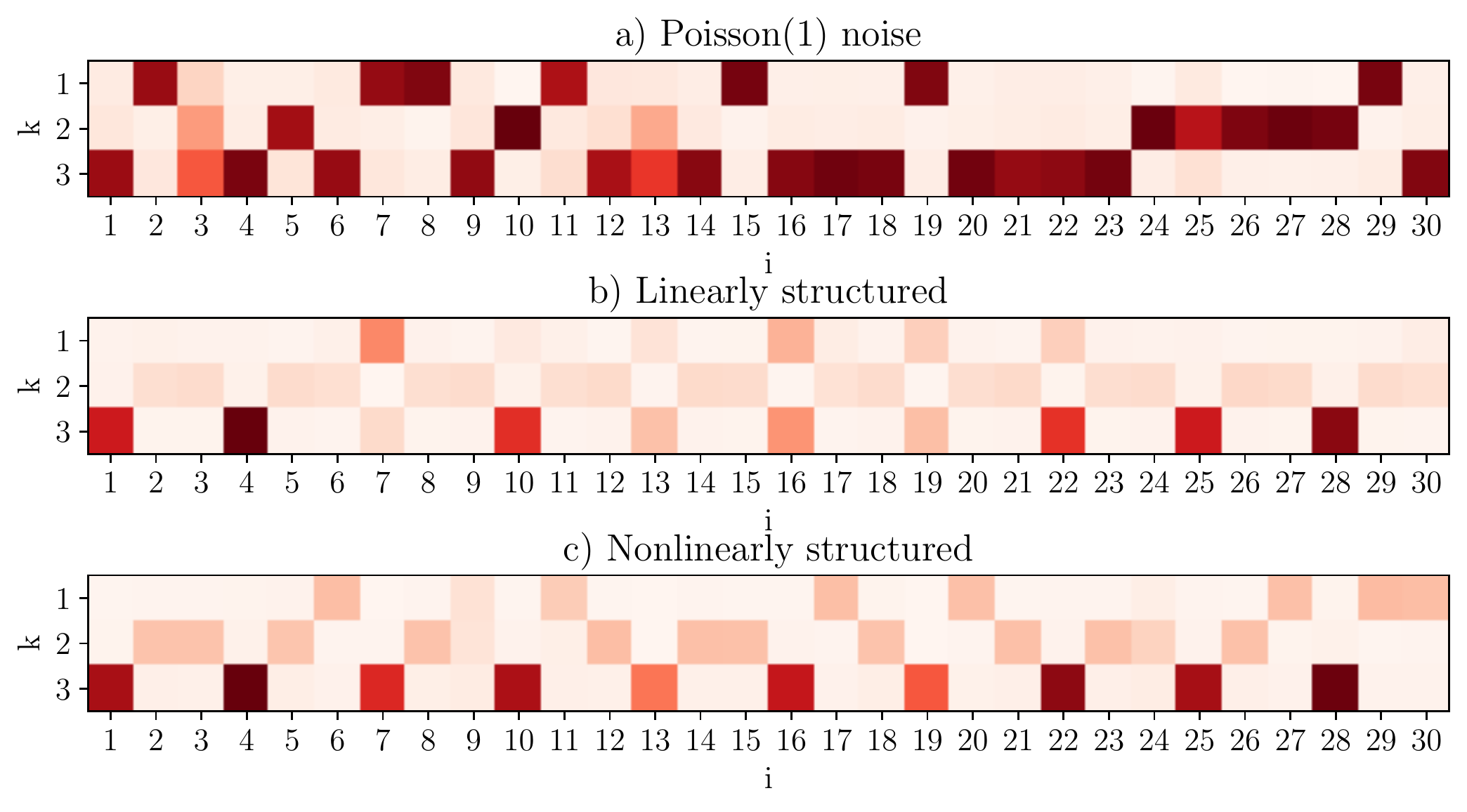}
    \caption{{Decoder matrices $\mathbf{B}=(\beta_{ki})$ for standard HPF} factorization of the synthetic datasets of Fig.~\ref{fig:simulation_factorization} using the python package \texttt{hpfrec}. Shown are posterior means.}
    \label{fig:hpfrec_sims}
\end{figure}

\textbf{Comparison to standard HPF.}
 In standard HPF~\citep{gopalanScalableRecommendationPoisson2014} only decoder matrices are inferred, and encoders are not explicitly reconstructed. 
Fig.~\ref{fig:hpfrec_sims} demonstrates factorization of the same synthetic datasets using standard hierarchical Poisson matrix factorization~\citep{gopalanScalableRecommendationPoisson2014} found in the \texttt{hpfrec} package. 
In all three examples, the standard HPF fails to remove independent noise items from the factor model.
In contrast, our method excludes all irrelevant features from the factor model (Fig.~\ref{fig:simulation_factorization}).
In this case, the encoder is not explicitly solved. It is incorrect to read the decoder matrix $\mathbf{B}$ row-wise, to say for instance that the first factor in Fig.~\ref{fig:hpfrec_sims}a is determined from items $\{2,3,7,\ldots, 29\}.$ 
However, results from standard HPF are often erroneously interpreted in this manner, suggesting that there is structure to the dataset even when none is present.
Additionally, for this reason, the generative process fails to adequately fit the data in that it correlates sources of independent noise.

\subsection{Comorbidities from billing codes}

\begin{figure}[ht]
    \centering
    \includegraphics[width=0.8\textwidth]{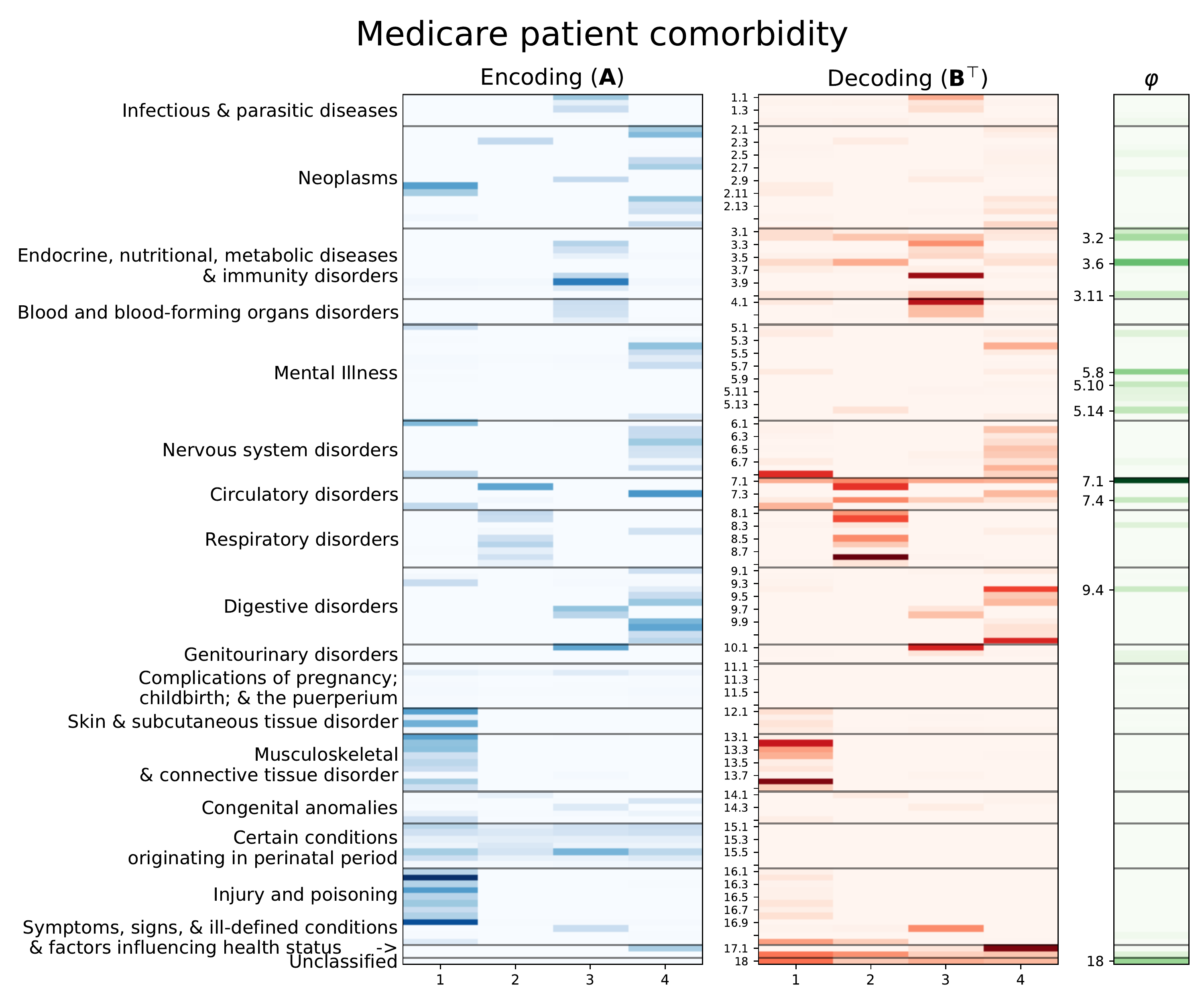}
    \caption{\textbf{Medicare comorbidity factorization} for inpatient visits based on medical claims from a 5\% sample of the Medicare Limited Dataset (LDS), in four factor dimensions. Prior to factorization, we mapped each raw ICD diagnostic code into the second tier of the Clinical Classification Software (CCS), counting the number of codes present within each broad category. Shown are posterior means. \textbf{Left:} encoding $\bA = (\alpha_{ik})$, \textbf{middle:} decoding $\mathbf{B}^\top = (\beta_{ki})^\top$,  \textbf{right:} background $\boldsymbol{\varphi} = ({\varphi}_i)$
    }
    \label{fig:Dx_factorization}
\end{figure}


As a real-world case study, we used a 5\% sample of the Medicare Limited Data Set (LDS) over the years 2009--2011 to discover a representation of inpatient comorbidity during a hospital visit.
The LDS consists of de-identified medical claims data for Medicare and Medicaid recipients across the United States.
Pursuant to a data use agreement, the Center for Medicare and Medicaid Services (CMS) provides a 5\% sample of this dataset for research purposes.
A single hospital visit consists of multiple claims across providers and types of services. 
No standard method for grouping claims into hospital visits within claims data exists. 
A heuristic algorithm (often called a \emph{grouper} algorithm) is used to reconstruct medical and billing events during a visit or type of service from claims. 
Our grouper algorithm collapsed the claims into $U=1,949,788$ presumptive inpatient visits. 
Diagnostic codes were then made coarser-grained by backing off from the original $\approx$ 13,000 ICD-9-CM to $136$ clinically-relevant categories using the top two levels of the CCS multilevel classification.
Within each visit, we counted the number of codes that fell into each of the CCS categories.
 Fig.~\ref{fig:Dx_factorization} presents the encoding matrix $\bA=(\alpha_{ik})$ for a factorization of comorbidities into four dimensions, the transpose of the decoding matrix $\mathbf{B}=(\beta_{ki})$, and the vector of background process rates $\boldsymbol{\varphi}=(\varphi_i)$ for the same model.

The values in the encoding matrix provide coefficients that are used in Eq.~\ref{eq:theta_constraint} to produce a weighted sum of billing code counts, which is then used to formulate a representation.
One may read the encoding matrix column-wise 
in order to determine the feature composition of each of the representation factors.
Being able to do so facilitates interpretation of the factor model, by allowing one to understand a single factor at a time by focusing on subsets of the original features.
For example,
conceptually, it is easy to see what factor 2 represents. 
It is computed by tallying up various billing codes that pertain to lung ailments -- lung cancer (CCS 2.3),  several broad respiratory disorders (CCS 8.x), and heart disease (CCS 7.2).
The relative weights of these codes are depicted by the color of the shading.

The interpretation of matrix factorization is not provided by the decoding matrix, which is the sole output of standard HPF.
The decoding matrix provides only an incomplete picture of the structure of the data.
Recall  from  Fig.~\ref{fig:interpretability} that the decoding matrix is not to be read row-wise (or column-wise in the transpose depiction of Fig.~\ref{fig:Dx_factorization}).
Looking solely at the decoding, and reading it in this incorrect way, one might erroneously conclude that lower respiratory disorders (CCS 8.8) are the main determinant of factor 2.
However, this conclusion is incorrect -- diagnoses of heart disease (CCS 7.2) are the main determinant.

The decoding matrix also provides misleading insights on feature selection.
For example, lung cancer (CCS 2.3) appears only very faintly in the decoding.
For this reason, one might come to the erroneous conclusion that it is not important as a feature.
However, recall that by Eq.~\ref{eq:hpf}, relatively rare features will only faintly register in the decoder matrix.
The rate of lung cancer diagnoses is low, yet lung cancer diagnoses are predictive of - and coincide with - other respiratory issues.
Hence, lung cancer (CCS 2.3) appears strongly in the encoder matrix.

After computing representations for the entire dataset, one may cluster patients into diagnostic groups. One way of doing so is through stratification, for instance into low, medium, and high groups for each of the four factors in Fig.~\ref{fig:Dx_factorization}.
Doing so based on quantiles yields thresholds between the groups, defined over representations.
Using Eq.~\ref{eq:reparameterization} one can easily convert these thresholds into decision rules on the counts. 
For example, a patient presenting with one or more lung cancer-related diagnoses would generally be placed in the medium or high strata for factor 2, depending on the number of other respiratory billing codes in that visit. 
Recall that the input data is sparse so in general the low strata for each representation would encompass people who had no or very few of the associated diagnoses.
As a first step in a modeling pipeline, one could use the strata to segment a larger overall model, so that the model has both local and global behavior, while making it easy to interpret how individual or collective billing codes contribute to an overall prediction.

\section{Discussion and Conclusion}

We introduced a constrained HPF where an encoder transformation is learned self-consistently with the matrix factorization. By imposing sparsity on the encoder, rather than on the decoder, we improve interpretability of HPF.
We demonstrated the approach on simulated data, showing that the method can successfully separate structure from noise, even when the model is mis-specified. We also presented a comorbidity factorization as a case study.

Although we focused on Poisson factorization,  our central argument holds for other sparse matrix factorization methods.
Sparse decoding matrices (loading matrices) inferred using these methods are generally not orthogonal.
Unlike in classical or orthogonally-rotated PCA,
the transpose of these decoding matrices does not correspond to their pseudo-inverse.
Hence, decoding matrices should never be interpreted row-wise (Fig.~\ref{fig:interpretability}c).

\subsection*{Limitations and extensions}

Our method relies on the horseshoe prior for sparsification.
The horseshoe prior relies on scaling hyperparameters, which control the effective sparsity of the method.
In order to make these priors scale asymptotically with data size~\citep{piironenSparsityInformationRegularization2017},
we chose to scale this prior using $\nicefrac{1}{\sqrt{UI}}.$
Empirically, this choice, along with the scaling of the priors on $\beta_{ki},$ led to desirable behavior in simulations like those in Fig.~\ref{fig:simulation_factorization} under several combinations of $U$ and $I$.
In effect, we have taken the liberty of formulating our method based on these considerations so that it is usable without needing to manually choose hyperparameters.
One may wish to rescale the horseshoe prior in order to control sparsity.
Further guidance to the regularization scale will require analysis outside of the scope of this manuscript.

We note that one could also formulate our method using the sparsifying priors found in \citet{gopalanScalableRecommendationPoisson2014}.
The chief advantage of the original HPF formulation is in how it yields explicit variational inference updates.
However, our method yields well to ADVI, achieving convergence with a learning rate of $0.05$ in approximately $100$ epochs in all included examples.

A limitation of our method, shared by standard HPF, is that a generalized linear model does not have the expressivity of nonlinear paradigms such as deep learning. 
For some applications, with sufficient data, nonlinear models may be more performant.
We note that one could place non-linearity in either $f_i$ or $g_i$, without compromising interpretability of the representation.
 These functions may be learned using Gaussian processes~\citep{changPathIntegralApproachBayesian2014}
  splines,
 or even neural networks, making the method more like other probabilistic autoencoders. 
 So long as the conditions of monotonicity and a fixed point at $y=0$ are maintained, the overall method remains interpretable. 
However, the simplicity of HPF offers statistical advantages that help it generalize better than deep learning except when there is enough data to learn any true nonlinearities in the true generating process. 
Additionally, while we do not explore this, a strength of Bayesian modeling is that it provides a principled approach to incorporating prior information. 
One could encourage or discourage features from co-factoring by setting suitable priors on the encoding and decoding matrices.

\subsubsection*{Acknowledgements}

We thank the Innovation Center of the Center for Medicare and Medicaid services for providing access to the CMS Limited Dataset.
We also thank the {med$\varepsilon_{\text{rrata}}$} team (particularly Joe Maisog) for their support in various data pre-processing tasks. JCC, AZ, and BD were supported in-part by the US Social Security Administration.
PF, JH, CCC, and SV were supported by the Intramural Research Program of the NIH, NIDDK. This work used the Extreme Science and Engineering Discovery Environment (XSEDE)~\citep{xsede}, which is supported by National Science Foundation grant number ACI-1548562  through allocation TG-DMS190042. We also thank Amazon Web Services for providing computational resources and \href{https://boostlabs.com/}{Boost Labs LLC} for helping with visualization.

\bibliography{irtvae}

\newpage
\clearpage
\appendix

\renewcommand{\thefigure}{S\arabic{figure}}
\renewcommand{\thesection}{S\arabic{section}}
\renewcommand{\thepage}{S\arabic{page}}
\setcounter{figure}{0}
\setcounter{section}{0}
\setcounter{page}{1}

Here we provide additional experiments. We note that the code for reproducing these experiments and those in the main manuscript can be found at \href{https://github.com/mederrata/spmf}{github:mederrata/spmf}. These supplemental examples can be found at \href{https://colab.research.google.com/drive/1RU-FSHqzU8y3GjhoFKVxQG_P9jOLr_Is?usp=sharing}{Google Collaboratory}. Please refer to the notebooks therein, where one can also find the details behind hyperparameters and optimization. In general, we did little tuning of the method beyond tuning the learning rate for stable inference.

\section{Comparing choices of $f, g$}
\begin{figure}
    \centering
    \includegraphics[width=\textwidth]{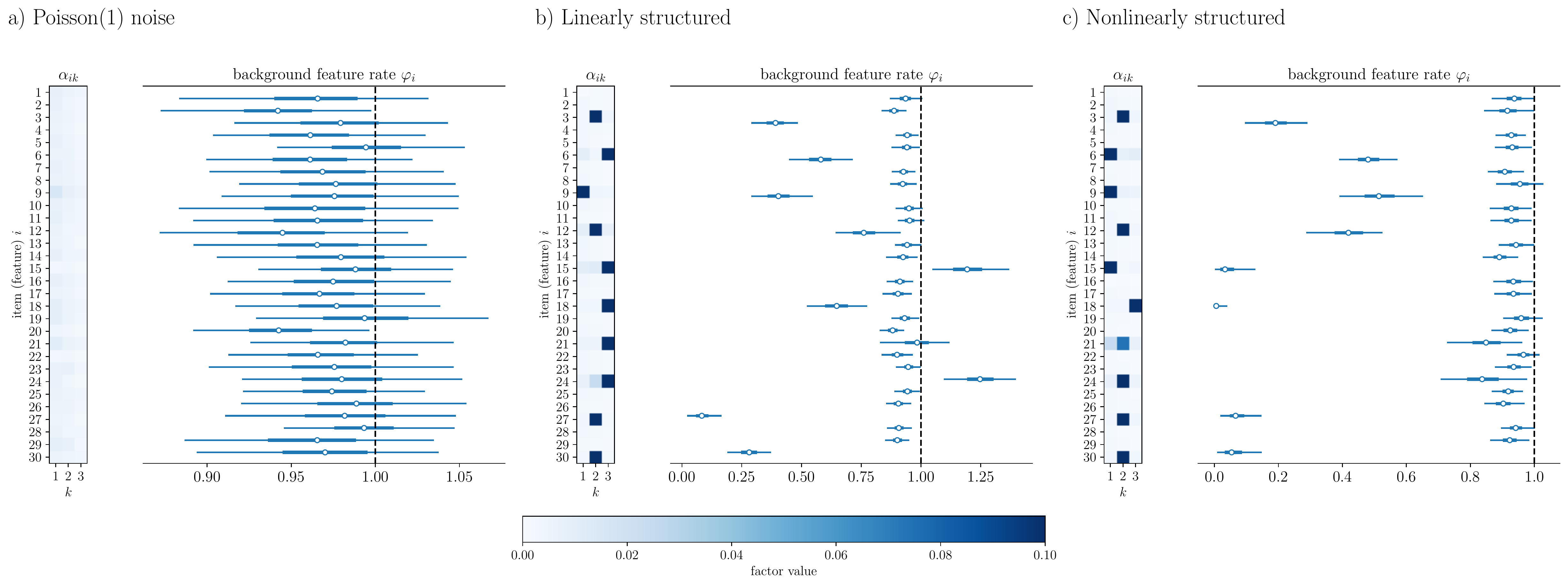}
    \caption{\textbf{Factorization based on the logarithmic link function} of Section~\ref{sec:model} of the synthetic dataset of Fig.~\ref{fig:simulation_factorization}}.
        \label{fig:log_sim_factorization}
\end{figure} 

In our method, we are free to choose functions $f_i, g_i$.
We evaluate models for predictive power without refitting by using the WAIC. 
Here we provide an example of factorizations under different functions $f, g$, and compare the models.
In Fig.~\ref{fig:log_sim_factorization}, we performed factorization of the synthetic datasets of Fig.~\ref{fig:simulation_factorization} using  logarithmic link function of Section~\ref{sec:model}. 
Although the model is mis-specified, the key structure of the data is still exposed and irrelevant features are removed. 

We then used WAIC to compare the use of the log link function versus the identity function.
On the basis of predictive accuracy, the two models are similar as shown in  Table~\ref{tab:waic_synthetic},
so the method is not sensitive to this choice.
\begin{table}
    \centering
    \begin{tabular}{c|c | c}
     & $f_i(x)=x/\eta_i$ & $f_i(x)=\log(x/\eta_i + 1)$ \\ \hline
        Poisson(1) noise & $(3.54 \pm 0.02) \times 10^5$ & $(3.54 \pm 0.02) \times 10^5$\\
       Linearly structured  & $(4.45 \pm 0.03) \times 10^5$ & $(4.43 \pm 0.03)\times 10^5$\\
       Nonlinearly structured &$(4.13 \pm 0.03) \times 10^5$ & $(4.13 \pm 0.03) \times 10^5$\\ \hline
    \end{tabular}
    \caption{\textbf{Model comparison using WAIC} ($\pm$ standard error) for factorizatons of the synthetic data. Lower is better.}
    \label{tab:waic_synthetic}
\end{table}

\section{Sample sizes}

\begin{figure}
    \centering
    \includegraphics[width=0.95\textwidth]{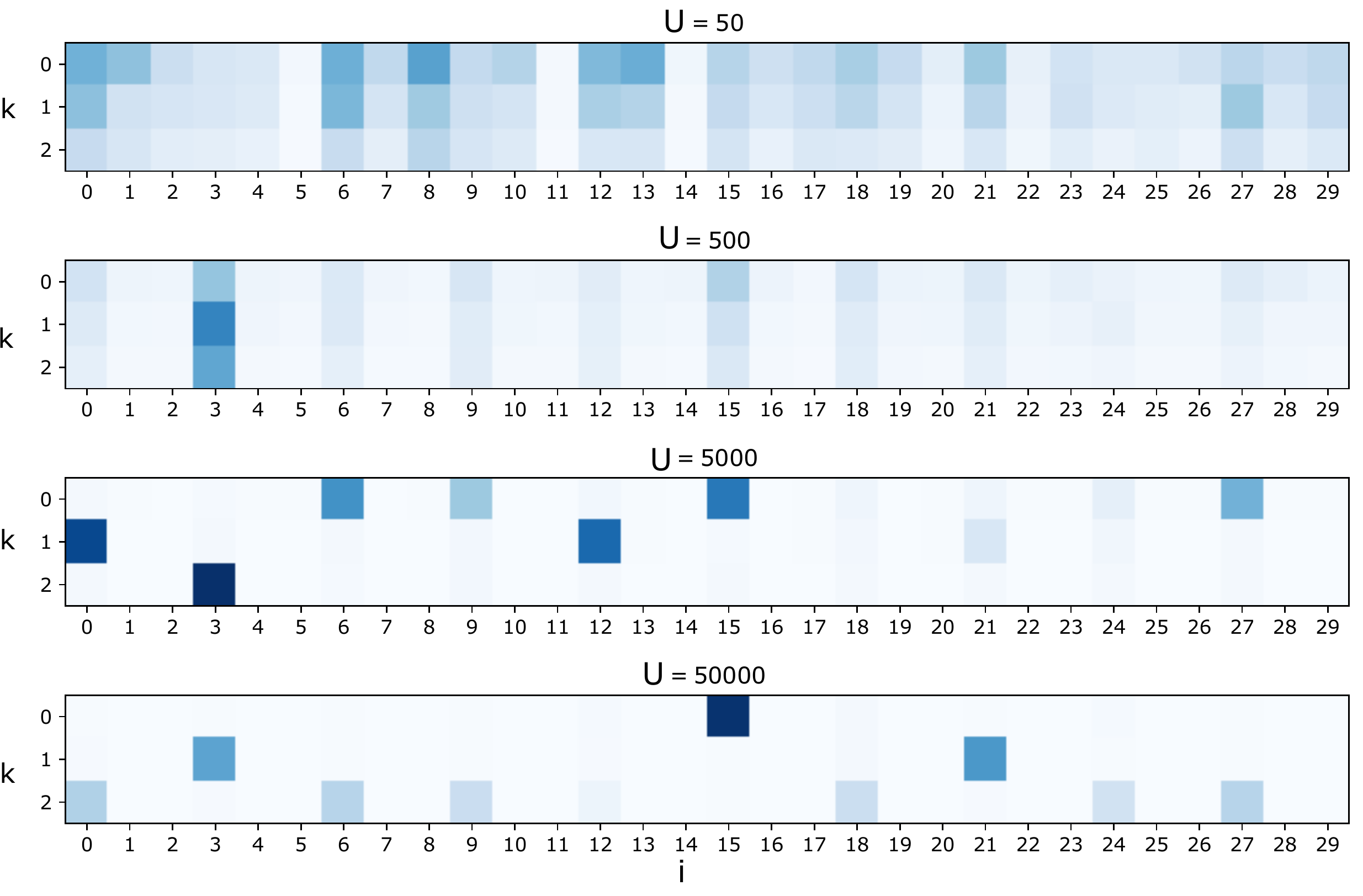}
    \caption{\textbf{Factorization under different data set sizes} of the nonlinearly generated data of Fig.~\ref{fig:simulation_factorization}}
    \label{fig:simulations_n}
\end{figure}

\begin{figure}
    \centering
    \includegraphics[width=0.95\textwidth]{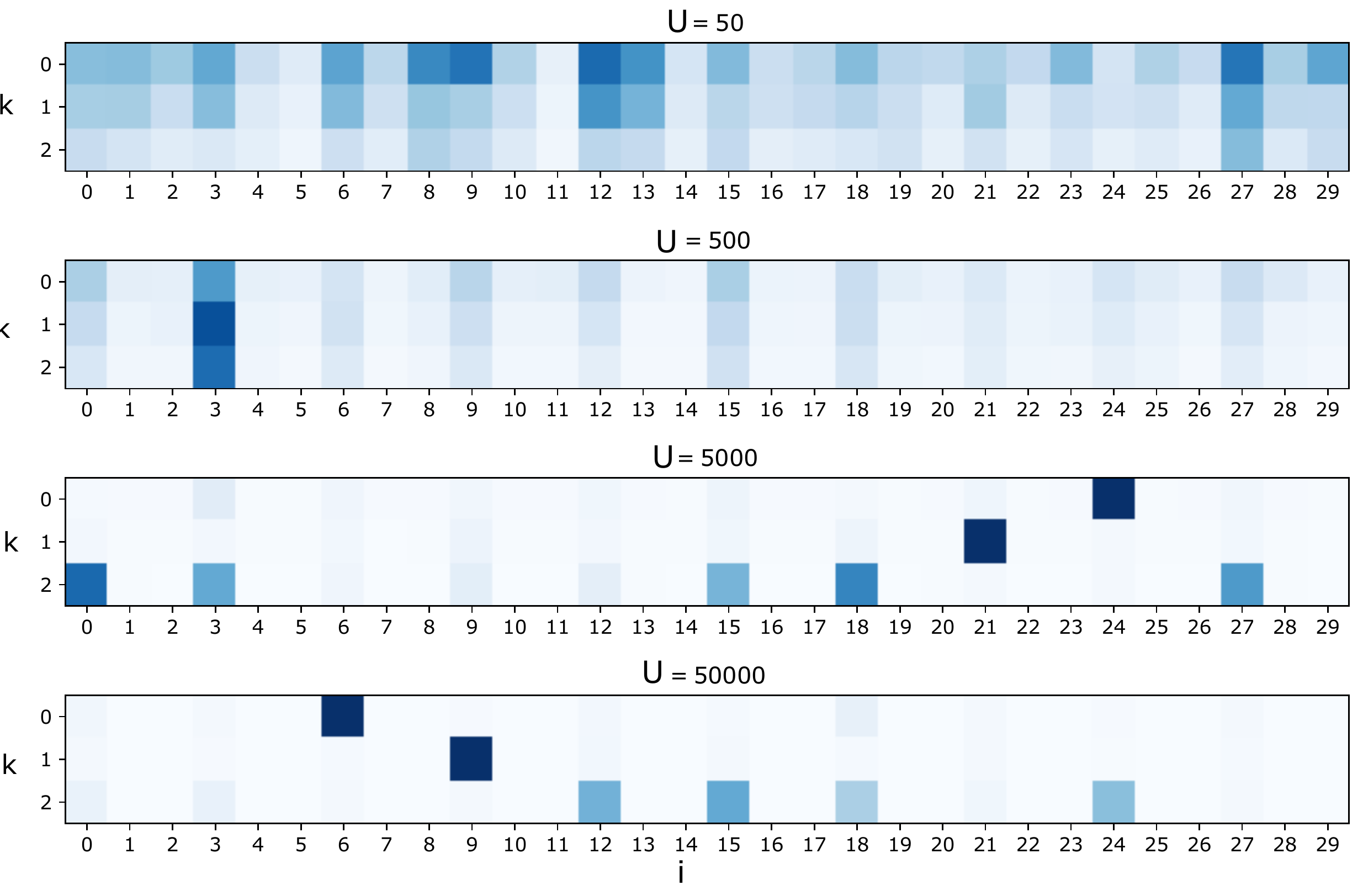}
    \caption{\textbf{Factorization under different data set sizes} of the nonlinearly generated data of Fig.~\ref{fig:simulation_factorization} using the logarithmic link function.}
    \label{fig:simulations_n_lt}
\end{figure}

For a systematic exploration of how sample size affects results, we used the nonlinearly generated synthetic dataset of Fig.~\ref{fig:simulation_factorization} and examined factorization as we varied $N$.
Fig.~\ref{fig:simulations_n} presents examples of these factorizations using the standard HPF link functions of Eq.~\ref{eq:standard_link} and Fig.~\ref{fig:simulations_n_lt} presents factorizations using the logarithmic link of Section ~\ref{sec:model}.

\begin{figure}
    \centering
    \includegraphics[width=\textwidth]{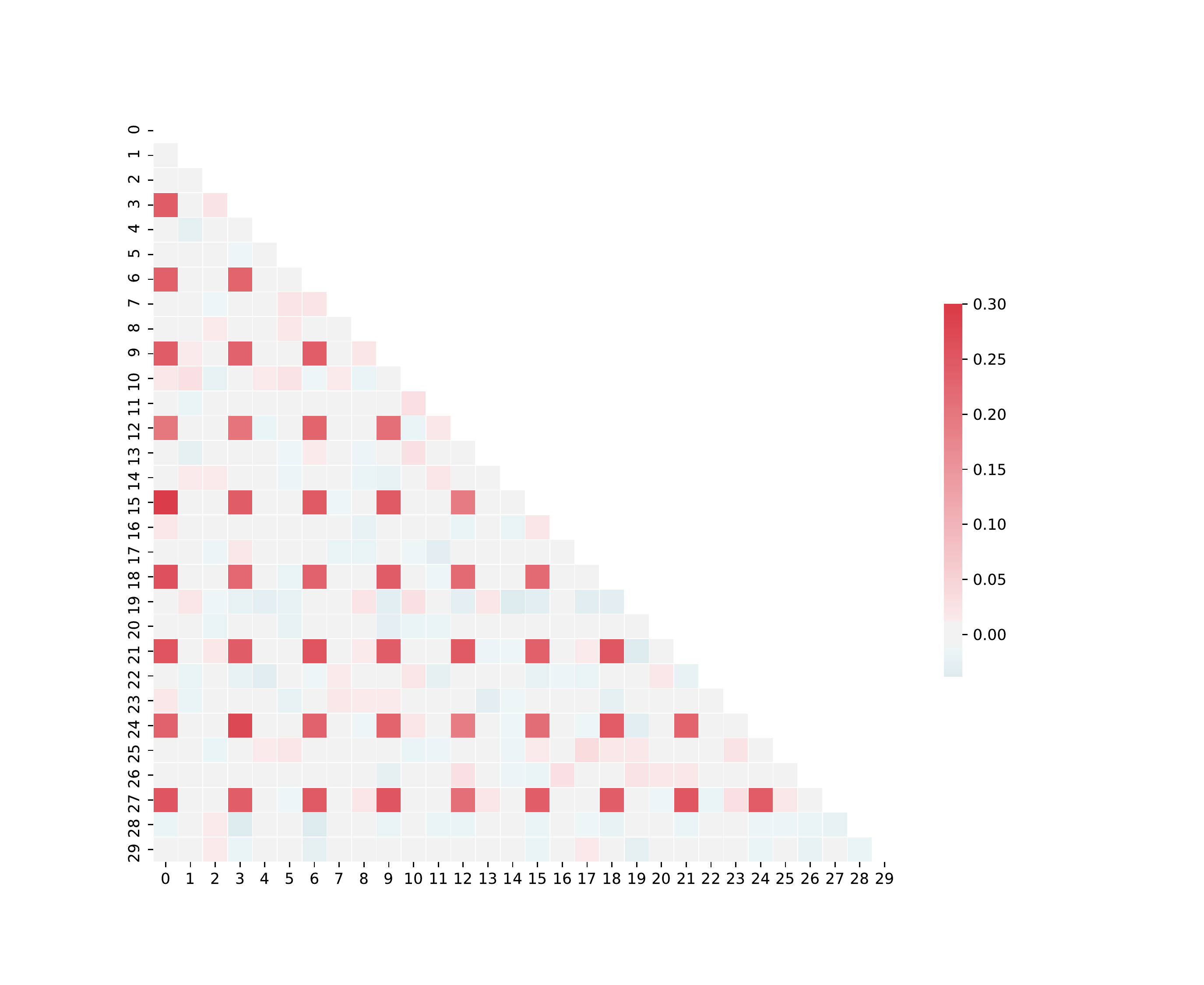}
    \caption{{Correlation of the nonlinear synthetic dataset features}}
    \label{fig:correlations}
\end{figure}

For $U$ sufficiently large, the factorizations successfully remove the irrelevant background features.
However, the structure of the factors is inconsistent as $U$ changes.
Examining the correlation matrix of this dataset (Fig.~\ref{fig:correlations}) sheds light on this behavior.
Since the true generating data for this example is dense in every third feature, these features are highly correlated.
Hence, without a spare substructure to select, the factorization settles on one of the many sparse approximations to the truly dense process.

\end{document}

%% file: math_commands.tex
\usepackage{amsmath,amsfonts,bm}

\def\eqref#1{equation~\ref{#1}}

\def\1{\bm{1}}

\DeclareMathAlphabet{\mathsfit}{\encodingdefault}{\sfdefault}{m}{sl}
\SetMathAlphabet{\mathsfit}{bold}{\encodingdefault}{\sfdefault}{bx}{n}

